\documentclass[review]{elsarticle}

\usepackage{lineno,hyperref}

\journal{arXiv}
\usepackage{amsmath,amsfonts}
\usepackage{amssymb}
\usepackage{latexsym}
\usepackage{algorithmic}
\usepackage{graphicx}
\usepackage{textcomp}
\usepackage{xcolor}
\usepackage{caption}
\usepackage{subcaption}
\usepackage{booktabs}
\usepackage[T1]{fontenc}
\usepackage{amsmath}
\usepackage{balance}
\usepackage{xurl,multirow}
\usepackage{soul}
\usepackage{microtype}
\usepackage[linesnumbered,ruled,vlined,noline,noend]{algorithm2e}

\usepackage{url}
\definecolor{newcolor}{rgb}{.8,.349,.1}

\usepackage{changepage}
\usepackage{framed}
\usepackage{latexsym}
\usepackage{lscape} 
\usepackage{rotating}
\usepackage{tabularx} 
\usepackage[linesnumbered,ruled,vlined,noline,noend]{algorithm2e}
\usepackage{soul} 
\usepackage{balance}
\usepackage{flushend}
\usepackage{microtype}

\newcommand\best[1]{{\textcolor{red}{\textbf{#1}}}}
\newcommand\x{\mathbf{x}}
\newcommand\y{\mathbf{y}}

\newcommand\w{\mathbf{w}}

\begin{document}

\begin{frontmatter}

\title{A Decidability-Based Loss Function}

\author[decomufop]{Pedro Silva \corref{mycorrespondingauthor}}
\ead[url]{http://www.decom.ufop.br/csilab/}
\ead{silvap@ufop.edu.br}

\author[decomufop]{Gladston Moreira}
\author[decomufop]{Vander Freitas}
\author[decomufop]{Rodrigo Silva}
\author[parana]{David Menotti}
\author[decomufop]{Eduardo Luz}

\cortext[mycorrespondingauthor]{Corresponding author}

\address[decomufop]{Computing Department, Federal University of Ouro Preto, Campus Morro do Cruzeiro, Ouro Preto-MG, Brazil}
\address[parana]{Department of Informatics, Federal University of Paran\'a, Centro Polit\'ecnico, Jardim das Am\'ericas, Curitiba-PR, Brazil}

\begin{abstract}
Nowadays, deep learning is the standard approach for a wide range of problems, including biometrics, such as face recognition and speech recognition, etc. 
Biometric problems often use deep learning models to extract features from images, also known as embeddings. 
Moreover, the loss function used during training strongly influences the quality of the generated embeddings.
In this work, a loss function based on the decidability index is proposed to improve the quality of embeddings for the verification routine.
Our proposal, the D-loss, avoids some Triplet-based loss disadvantages such as the use of hard samples and tricky parameter tuning, which can lead to slow convergence.
The proposed approach is compared against the Softmax (cross-entropy), Triplets Soft-Hard, and the Multi Similarity losses in four different benchmarks: MNIST, Fashion-MNIST, CIFAR10 and CASIA-IrisV4.
The achieved results show the efficacy of the proposal when compared to other popular metrics in the literature. The D-loss computation, besides being simple, non-parametric and easy to implement, favors both the inter-class and intra-class scenarios.
\end{abstract}


\end{frontmatter}


\section{Introduction}

Deep learning disrupted the computer vision field, especially regarding feature extraction and image representation. In this context, convolutional neural networks (CNNs) play a significant role~\citep{krizhevsky2009learning,lecun2015deep}. Once deep CNNs are trained on a massive amount of data, it becomes a candidate to be used as a deep feature descriptor, not only in the domain or task in which the CNN was trained but in every computer vision problem, thanks to transfer learning techniques~\citep{goodfellow2016deep}. However, the loss function in which these CNNs were originally pre-trained, strongly influences the extracted deep representation features.
The loss function is of paramount importance for training a CNN model and several approaches have been proposed in the literature \citep{sharif2014cnn,schroff2015facenet,parkhi2015deep,wang2017normface,wang2019ranked,wang2019multi}. Among them, one very popular is the cross-entropy loss~\citep{goodfellow2016deep}, also known as softmax-loss since it is often preceded by a softmax operation.  Although originally designed for classification problems, it has been very successful in learning deep representative features along with CNN models~\citep{krizhevsky2009learning,sharif2014cnn}.

The softmax loss is widely used for several reasons, namely it is easy to interpret and implement, fast convergence, and for working well with different batch sizes (large or very small). Even though it is not designed for learning feature representations, the learned features are powerful enough for many tasks, such as face recognition~\citep{parkhi2015deep,wang2017normface}, ocular recognition \citep{luz2018deep,silva2018multimodal}, among others. However, in tasks in which it is necessary to know how close or how far the samples are concerning each other (such as the biometric verification task) on high-dimensional spaces, this type of loss is not the best option. 
There are efforts in the literature to adapt the softmax loss~\citep{ranjan2017l2,wang2017normface} to the task, but contrastive and triplet-based losses are still the ones that offer the best gains.

Contrastive loss better captures the relationship between two samples projected in a space (Euclidean, for example), by penalizing, during learning, negative samples (from impostors) and rewarding samples from the same (genuine) category~\citep{chopra2005learning}. Likewise, triplet-based loss explores the concept of similarities and dissimilarities between samples in a space, adding anchor elements.  There are many triplet-based losses in the literature, such as triplet-center loss \citep{he2018triplet}, quadruplet loss \citep{law2013quadruplet} and in general, triplet-based loss produce better results, overcoming other pair-wise losses such N-pairs loss~\citep{sohn2016improved}, binomial deviance loss \citep{yi2014deep}, histogram loss \citep{ustinova2016learning} and Multi-Similarity Loss
~\citep{wang2019multi}. 

Low convergence represents a major problem for triplet-based losses. Besides, given a set of samples, it is not trivial to find positive or negative instances to use as hard pairs, nor is it easy to fine-tune the margin that separates them~\citep{parkhi2015deep}. Notwithstanding, tripled-based losses are still the most popular losses in the literature, despite their limitations.

In this paper, we propose a new loss function, the D-loss, based on the decidability index~\citep{daugman2000biometric}.
Daugman \cite{daugman2000biometric} highlights this index as a quality measure for biometric systems, which is often used in the literature for this purpose~\citep{de2018insights, luz2018deep}. 
The D-loss assures both inter-class and intra-class separability, and, unlike triplet-loss, avoids the difficult problem of finding hard positive and negatives samples. It also provides better convergence, since it does not require parameter adjustment, and is easy to implement. 

The contributions of this paper can be summarized as follows:\vspace{-0.2cm}
\begin{itemize}
    \item A new loss function, the D-Loss, based on the decidability index, which is intuitive and easy to compute and implement. The D-loss is also suitable for training models which aim at data representation and, unlike triplets loss, it favors both inter-class separability and intra-class approximation.\vspace{-0.2cm}
    \item Under the same conditions, the D-loss overcomes three other popular loss functions (Triplets Soft-hard Loss, Multi-Similarity Loss, and Softmax-Loss) in MNIST-FASHION, CIFAR-10 and presents comparable results for MNIST. Therefore, a competitive loss function.\vspace{-0.2cm}
    \item The D-loss has converged well on higher capacity networks such as the networks of the EfficientNet family. Results with D-loss overcome all the three other popular loss functions (Triplets Soft-hard Loss, Multi-Similarity Loss, and Softmax-Loss) evaluated here, for the ocular database CASIA-IrisV4 on the same model (B0) and training conditions.
    
\end{itemize}

The manuscript is organized as follows: Section 2 presents a background based on related works. Section 3 describes the methodology, followed by the experiments and discussion in Section 4. Finally, the conclusion appears in Section 5.

\section{Background and Related Works}
\label{sec:related}

In this section, two categories of loss functions are introduced: one designed for classification tasks and another for verification tasks, usually carried out in a pairwise fashion.

\subsection{Classification Losses -- Softmax Loss}

\begin{figure}[!htb]
    \centering
    \includegraphics[width=.97\linewidth,page=1,trim={0cm 0cm 0cm 0cm},clip]{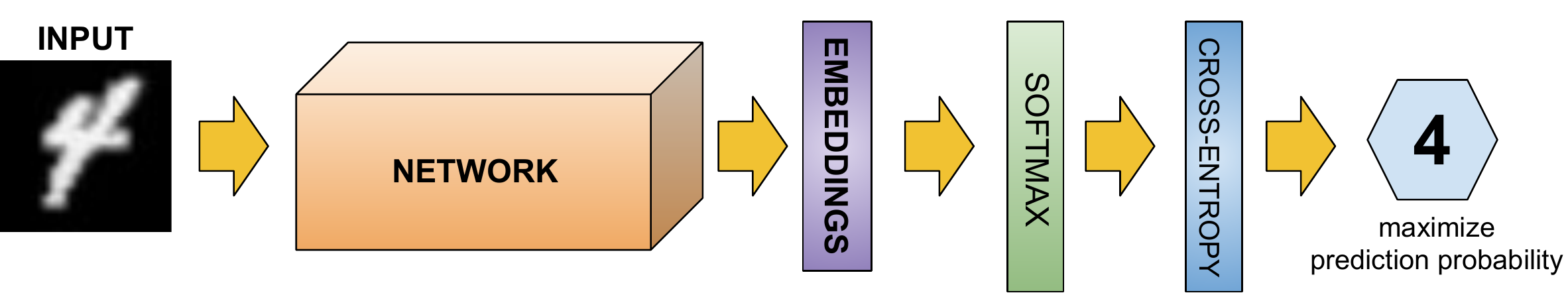}
    \vspace{-0.6em}
    \caption{Cross-entropy loss applied over a Softmax activation. To turn easier its reference in text, the name ``Softmax loss'' is used.}
    \label{fig:softmax_diag}
\end{figure}

Given a training set $\mathcal{X} = \{ \x_1,\x_2,...,\x_N\}, \x_i \in \mathbb{R}^{m \times n}$, and their respective labels $\mathcal{Y} = \{ \y_1,\y_2,...,\y_N\}, \y_i \in \{1,...,C\}$, where $C$ is the number of classes in the problem, the softmax loss function is given by:

\begin{equation}\label{eq:softmax}
    \mathcal{L}_{Softmax} = - \sum_{i=1}^{N}\log \left ( \frac{\mathrm{e}^{ \left ( \w_{y_i}^T f(\x_i) + b_{y_i} \right )} }{\displaystyle\sum^C_{j=1} \mathrm{e}^{\left (\w_{j}^T f(\x_i) + b_{j} \right )}} \right )
\end{equation}

\noindent in which $\mathcal{L}_{Softmax}: \mathbb{R}^{m \times n} \rightarrow \mathbb{R}^{K \times C}$ is the learned feature map, or embedding, $K$ is the dimension of the embedding. The $w_j$ and $b_j$,  $j \in \{1,...,C\}$,  are the weights and biases of the last fully layer, respectively (see Figure~\ref{fig:softmax_diag}). This formulation enforces features to have larger magnitudes and a radial distribution~\citep{wang2017normface}. Thus, it does not optimize the features to have small dissimilarity scores for positive pairs or higher dissimilarity scores for the negative ones. Other approaches tried to mitigate such effects on softmax-loss.

\subsection{Pair-wise losses}

\subsubsection{Triplet losses}

\begin{figure}[!htb]
    \centering
    \includegraphics[width=.9\linewidth,page=1]{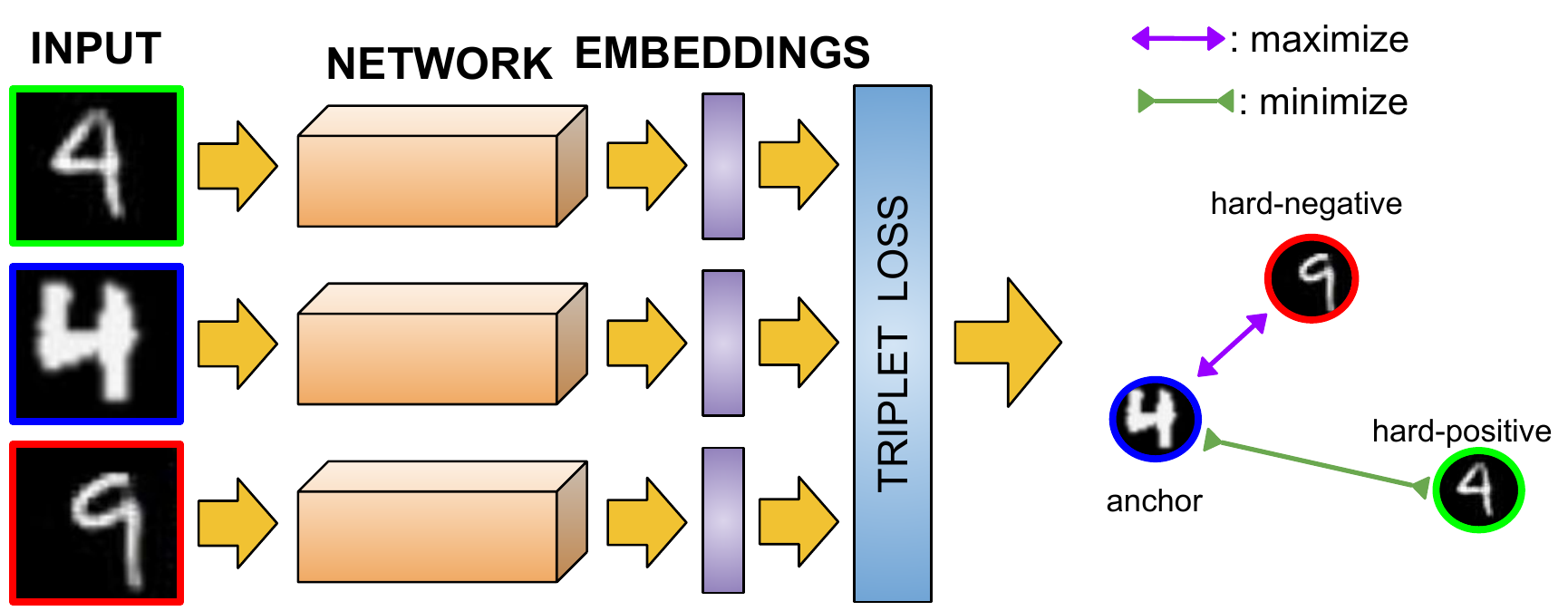}
    \vspace{-0.6em}
    \caption{Triplet loss optimization schema. The same network is applied to all instances.}
    \label{fig:triplets}
\end{figure}

A triplet is a set of three components: an anchor $x_i^a$, a sample of the same class $x_i^p$ (positive), and one from another class $x_i^n$ (negative) \citep{parkhi2015deep,schroff2015facenet,wang2019ranked}. The positive and negative pairs share the same anchor, and the aim is to make the embedding $f$ of the positive pair get closer, and the negative to separate according to

\begin{equation}
\label{eq:triplet_ineq}
\left \| f(x_i^a) - f(x_i^p) \right \|_2^2 + \alpha < \left \| f(x_i^a) - f(x_i^n) \right \|_2^2, 
\end{equation}
\noindent in which $\alpha$ is a desired margin, and $\left \| . \right \|_2^2$ is the squared L2 distance.
The triplet loss is given by 
\begin{equation}
    \label{eq:triplets}
    \small
    \mathcal{L}_{Triplet} = \sum_i^N \max\{
        0,
        \left \| f(x_i^a) - f(x_i^p) \right \|_2^2 -
           \left \| f(x_i^a) - f(x_i^n) \right \|_2^2 + \alpha \},
\end{equation}
\noindent for a set of $N$ triplets. Nonzero values appear when the inequality \eqref{eq:triplet_ineq} does not hold for a given margin $\alpha$. Figure \ref{fig:triplets} exhibits a sketch of the process, with the green arrow indicating the distance between a positive pair and in violet the distance between a negative.

As in other pair-based losses, sampling is an important step \citep{wang2019multi}. The dataset must have sufficient recordings of the same class to use as positive pairs, otherwise the triplets are not viable. Moreover, randomly chosen $x_i^p$ and $x_i^n$ easily satisfies the condition \eqref{eq:triplet_ineq} since the euclidean distances between the encodings of images of different people are likely to be large whilst those from the same person tends to be small. The main drawback of the random strategy is low convergence due to uninformative samples with negligible contributions \citep{wang2019multi}. 

One solution is to perform hard mining by choosing $x_i^p$ in a manner that $\text{argmax}_{x_i^p} \left \| f(x_i^a) - f(x_i^p) \right \|_2^2$, and $x_i^n$ with $\text{argmin}_{x_i^n} \left \| f(x_i^a) - f(x_i^n) \right \|_2^2$. This gives positive (negative) pairs whose distances to the anchor are the highest (smallest). In this case it is possible that $\left \| f(x_i^a) - f(x_i^n) \right \|_2^2 \approx \left \| f(x_i^a) - f(x_i^p) \right \|_2^2$ and the margin starts to play an important role.

The hard mining process demands high computational power and may lead to biased positive and negative pictures ranging from mislabeled to poorly imaged samples. Some strategies to overcome these issues are: (i) to generate triplets offline every $n$ steps and compute $\text{argmin}$ and $\text{argmax}$ on a subset of the data; (ii) generate triplets online seeking hard pairs from the mini-batch; (iii) semi-hard negative samples \citep{schroff2015facenet}.  

In some problems, images from the same class may present high dissimilarity when compared to images from different classes due to differences in lighting, color, and pose. The Ranked List Loss (RLL) \citep{wang2019ranked} and the Online Soft Mining (OSM) \citep{Wang2019softmining} preserve intra-class data distribution instead of shrinking the encodings into a single point.

The Triplet-Center Loss (TCL) \citep{he2018triplet} replaces the hard negatives with the nearest negative centers from the center loss. Other mining strategies explore only moderate positive pairs~\citep{shi2016miningPositive}, or consider different depths of similarity with sub-networks \citep{Yuan2017hdc}.

\subsubsection{Multi-Similarity Loss}

There are other pair-based loss functions, such N-pairs loss~\citep{sohn2016improved}, binomial deviance loss \citep{yi2014deep}, histogram loss \citep{ustinova2016learning} and Multi-Similarity Loss~\citep{wang2019multi}. 
Among them, outstanding results have been reported with Multi-Similarity Loss. Therefore, we consider the Multi-Similarity Loss in this work for comparison purposes..

The Multi-Similarity Loss is based on the pair weighting formulation, which analyzes simultaneously three types of similarities before making a decision: self-similarity, negative relative similarity, and positive relative similarity. The approach consists of a two-step scheme: hard pairs selection (hard mining) and weighting. First, pairs are selected by means of positive relative similarity, then the selected pairs are weighted using both self-similarity and negative relative similarity, inspired by binomial deviance loss~\citep{yi2014deep} and lifted structure loss~\citep{oh2016deep}. A framework is proposed to integrate the two steps, the General Pair Weighting (GPW) framework.

\section{Methodology}
\label{sec:method}

In this section, the decidability metric is formally described along with the training and optimization strategies.

We do not intend to compare our results with state-of-the-art metric learning methods, but rather to evaluate the robustness and discriminative potential of the representations obtained with the D-loss. We are especially interested in assessing D-loss performance in a biometric-like problem.

\subsection{Decidability}

\begin{figure}[!htb]
    \centering
    \includegraphics[width=0.8\linewidth]{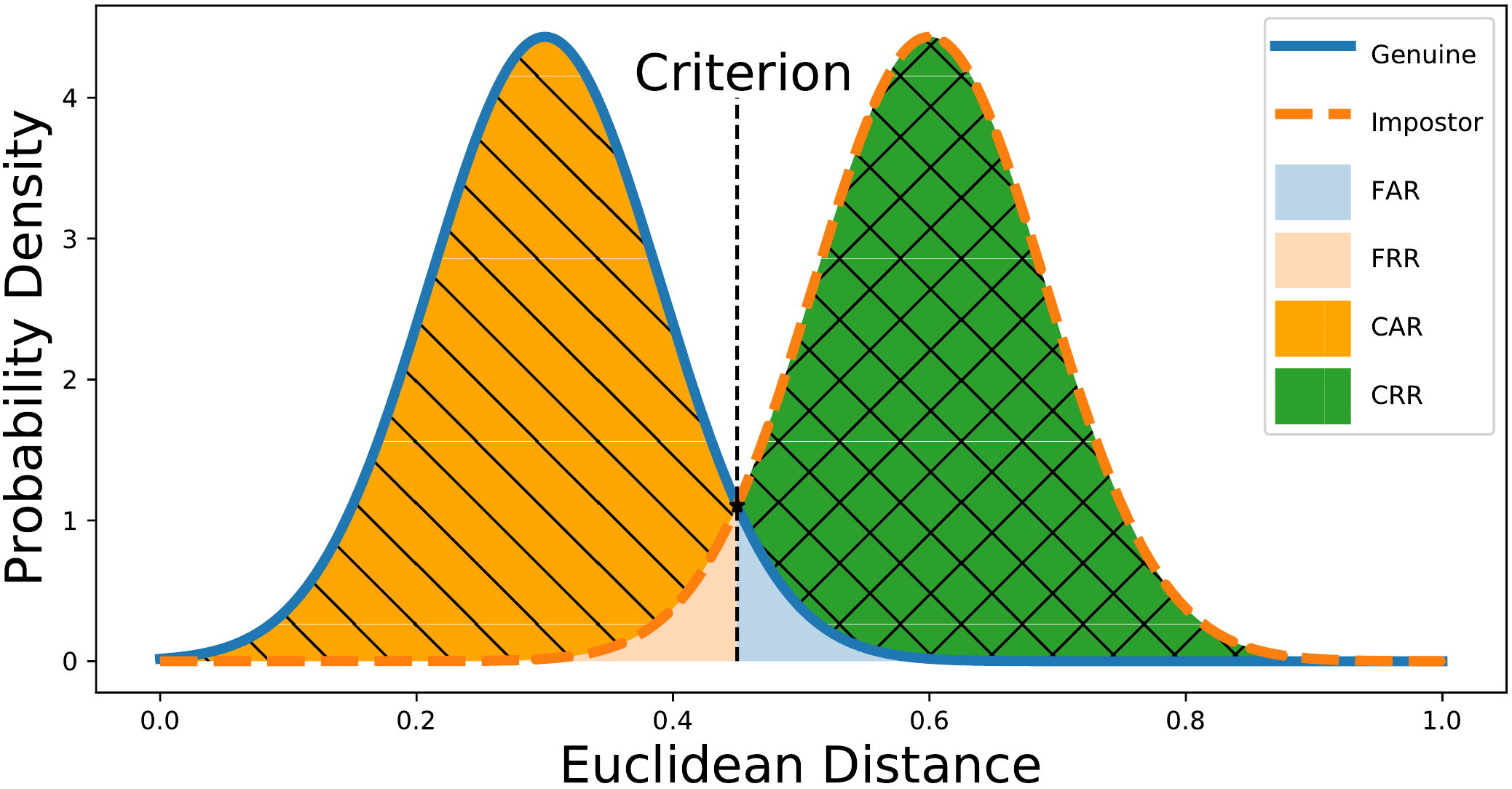}
    \vspace{-0.6em}
    \caption{Genuine and Impostor distributions. The Euclidean Distance is used to calculate the dissimilarity between two pairs. If the Euclidean Distance calculated is greater than a criterion, then, the individual is rejected, otherwise, it is accepted.}
    \label{fig:dists}
\end{figure}

A typical recognition system, such as biometric recognition, can be analyzed from four different perspectives: (i) False Accept Rate (FAR) in which an impostor is accepted as genuine, (ii) False Reject Rate (FRR) in which an genuine individual is classified as an impostor, (iii) Correct Accept Rate (CAR) in which an genuine individual is accepted, and (iv) Correct Reject Rate (CRR) in which an impostor is correctly not accepted.
The FAR and FRR are related to a system error and they are called Type I and Type II error respectively.
It is worth mentioning that we discuss the distance function from a point of view of dissimilarity  in this work.
Figure~\ref{fig:dists} shows the relation of the four perspectives when analyzed using the distribution of the scores.

\begin{figure}[!htb]
    \centering
    \includegraphics[width=0.9\linewidth]{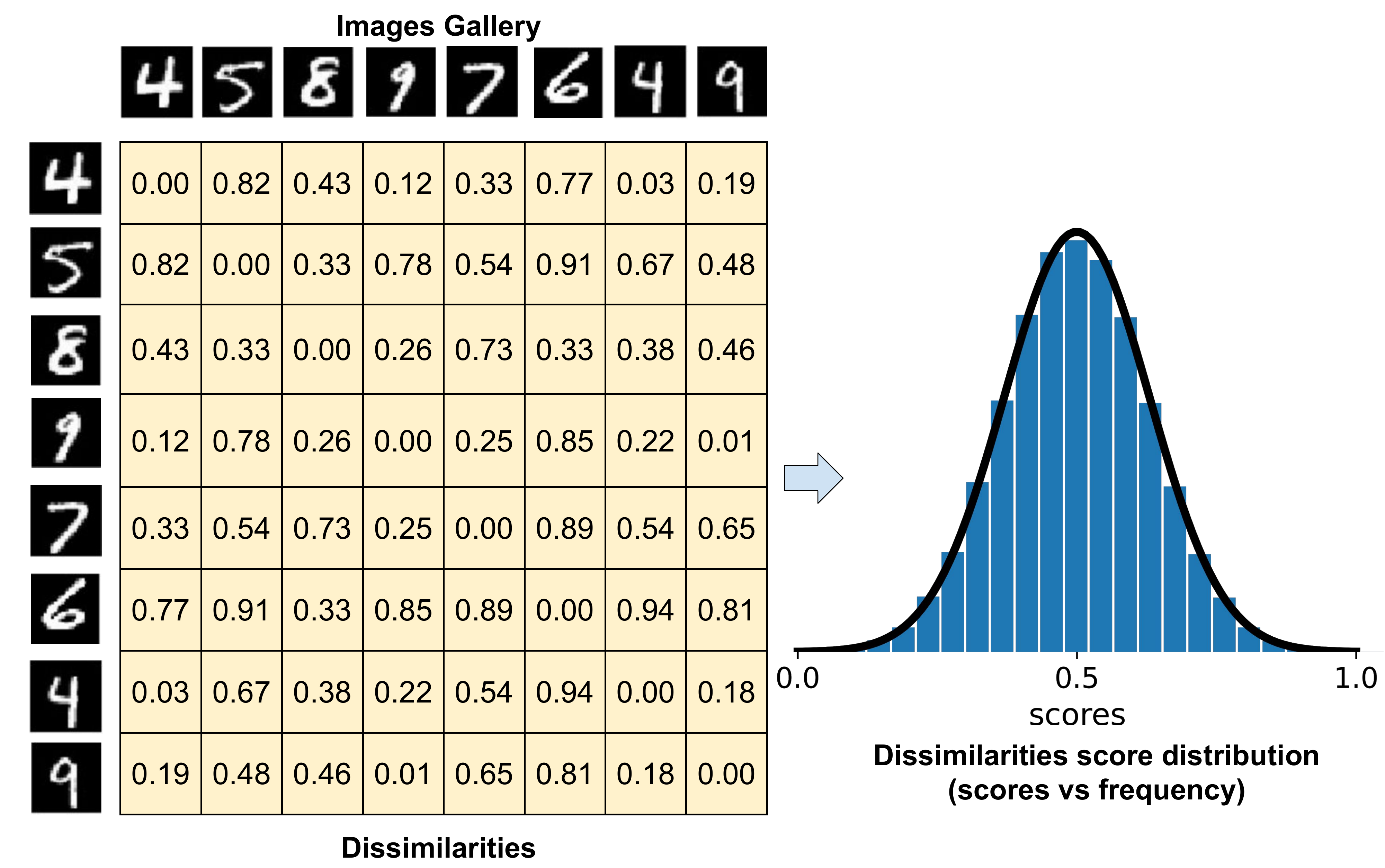}
    \vspace{-0.6em}
    \caption{Process of construction of a Histogram based on images dissimilarities.}
    \label{fig:hist}
\end{figure}

To illustrate, let $X$ be a gallery of $n$ samples. After propagating $X$ through a neural network, one obtains $S=f(X)$, in which $f$ is the embedding function, and $S$ is a representation in the embedding space. The dissimilarity score curves (genuine and impostor) use $S$ as input. The scores are generated by computing the distance of pairs of samples, on an all against all fashion, as shown in Figure
~\ref{fig:hist}. The scores are further mapped into categories (or bins), and a histogram of the frequencies is generated by
\begin{equation}
     \label{eq:freq_bin}
    F_{i} = \sum_{j} S_{j}
\end{equation}
\noindent where $F_{i}$ is the frequency count for a specific bin $i$, $j$ the number of samples that meets the bin conditions and $S_{j}$ the score related to the distance between a pair of samples.

Let $P_{G}$ and $P_{I}$ be the two distribution curves shown in Figure~\ref{fig:dists} regarding genuine and impostor, respectively. The $P_{G}$ is the probability density function of the dissimilarity scores, computed from the distance (through some distance metric) between two instances of the same class. Likewise, the $P_{I}$ curve represents the probability density of the dissimilarity scores, computed from the distance between instances of different classes.  

Thus, the four areas under the two distribution curves, such as illustrated in Figure~\ref{fig:dists}, express the probabilities of each decision metric (FAR, CAR, FRR, CRR).

The overlay of the authentic and the impostor scores is expected in a real scenario.
The criterion threshold (C) can be manipulated and the FAR, FRR, CAR, and CRR updated according to users' needs.
By manipulating this criterion and plotting the results in function of CAR and FAR, whose sum should result in one, a Receiver Operating Characteristic (ROC) or Neyman-Pearson curve can be created.

Ideally, the decision curve should be positioned as close as possible to the top left corner of the decision strategy curve in which the Correct Accept is close to 100\% and the False Accept (error), close to 0\%. However, this does not happen in the real world. To overcome those issues, arises the decidability index.

According to Daugman~\cite{daugman2000biometric}, the decidability along with the distribution curves are good descriptors for the decision curve and can be better than FAR and FRR to assess pattern recognition systems performance.

The decidability can be defined as a correlation of genuine and impostor scores as defined in
\begin{equation}
     \label{eq:dec}
    Decidability = \frac{|\mu_{I} - \mu_{G}|}{\sqrt{\frac{1}{2} (\sigma^{2}_{I} + \sigma^{2}_{G})}},
\end{equation}

\noindent in which $\mu_{G}$ and $\mu_{I}$ are the means of the two distributions (histogram curve of genuine and impostor scores) and $\sigma^{2}_{I}$ and $\sigma^{2}_{G}$ are their corresponding standard deviations. The decidability indicates how far the genuine distribution scores curve is from the impostor and the overlap between these two curves (as shown in Figure~\ref{fig:dists}).

Although the decidability index, as well as the other metrics presented in this subsection, is widely used in the biometric field, one could expand it to most pattern recognition problems. Since the decidability is independent of decision thresholds and quantifies, in a single scalar, the distance and overlap among two distribution curves, one hypothesis is that it might work as a loss function for training models aiming at image/signal representation.

\subsection{D-loss}

Many computer vision problems use deep learning models for data representation. Several authors use the cross-entropy loss (softmax-loss) to fine-tune models to a new domain or task, and it is also common to train models from scratch. To calculate the loss in a common supervised learning problem, with the cross-entropy loss, one uses the softmax operation to drive the output to probabilities (of each class), and the loss function aims to raise the accuracy on training data, according to a class hot encoded array. Thus, the problem is reduced to a classification problem, in with the main goal is not focused on the generation of better representation for the inputs, but on correctly classifying classes.

The state-of-the-art losses, triple-based ones, employ the concept of the anchor. Given an anchor sample, triplet-based loss aims to learn an embedding space, where positive pairs are forced to be closer to the anchor than the negative, by a margin.
The pair-based losses, such as constrative~\citep{chopra2005learning}, triplet-center~\citep{he2018triplet}, and quadruplet~\citep{law2013quadruplet}, put the focus on the generation of embeddings.

Ideally, those pairs tend to bring more information and enhance the power of the model to represent the data.

The pair selection represents a major issue for that kind of loss, since finding hard positive or negative samples related to anchor is not a trivial task. They also suffer from low convergence, and adjusting the margin parameter is not trivial~\citep{parkhi2015deep}.

\begin{figure}[!htb]
    \centering
    \includegraphics[width=.9\linewidth,page=1]{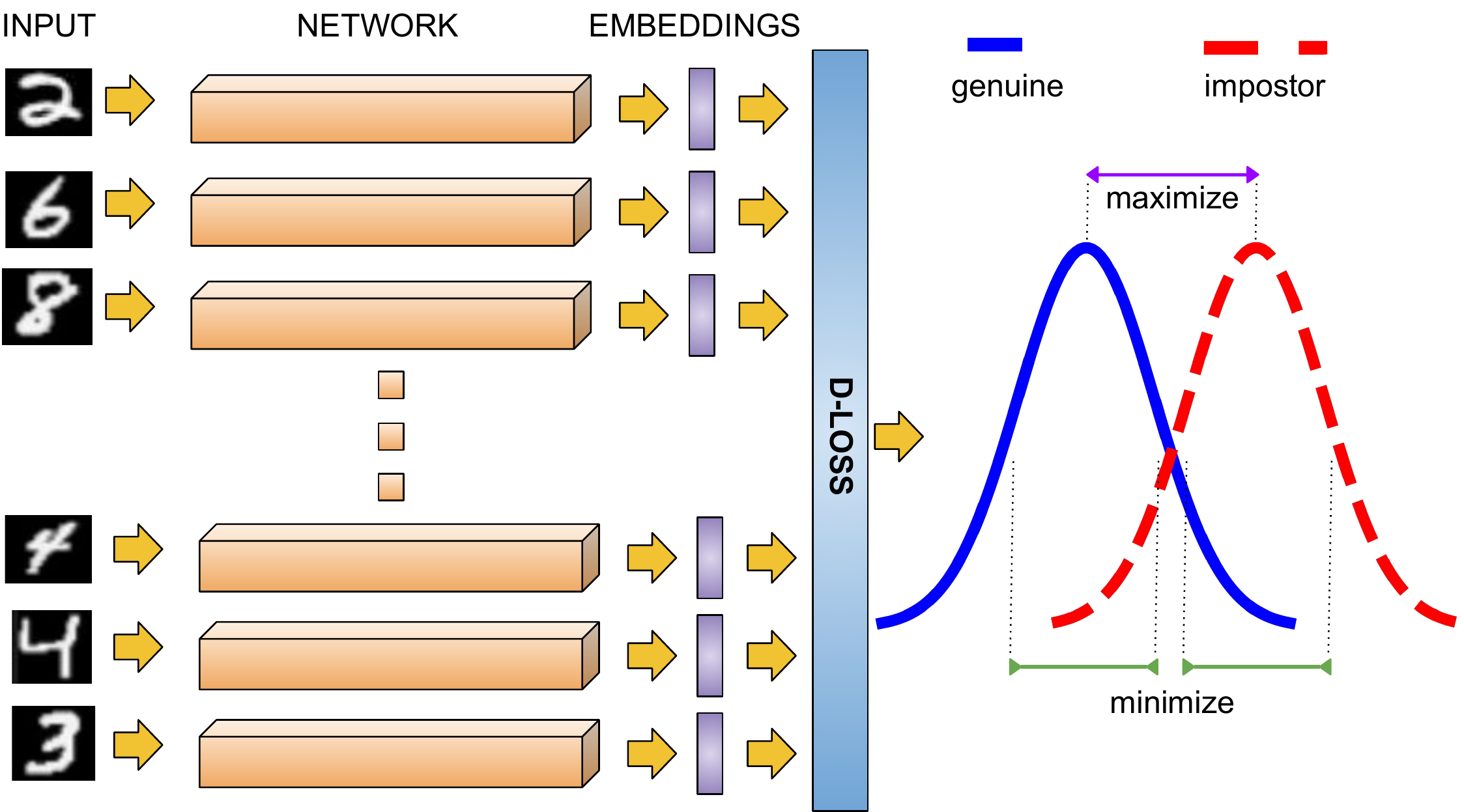}
    \vspace{-0.6em}
    \caption{D-loss optimization schema.}
    \label{fig:dloss}
\end{figure}

We propose a new loss function, the D-loss. Differently from others, it makes use of all samples in a batch and does not rely on the concept of an anchor. The optimization of Eq.~\eqref{eq:dec} separates $\mu_G$ and $\mu_I$ and reduces the variances $\sigma_G$ and $\sigma_I$ at the same time, thus improving both intra-class and inter-class scenarios as shown in Figure~\ref{fig:dloss}.

The decidability index can rise to infinity, and the higher its value, the best is the separation between the impostor and genuine distributions. To better suit a minimization objective, we define D-loss as follows:

\begin{equation}
    \label{eq:dloss}
    \mathcal{L}_{D\mbox{-}loss}(X, f) = \frac{1}{Decidability}
\end{equation}

\noindent
with $Decidability$ as defined in Eq.~\eqref{eq:dec}, and the embedding function $f$ based on Euclidean Distance
\begin{equation}
    \label{eq:euclidean}
    d_{i,j} = || f(x_i), f(x_j)||_2.
\end{equation}

The $X$ is given by $\{(x_i, y_i)\}^N_{i=1}$, in which $N$ is the number of samples, $x_i$ is the $i^{th}$ training data and $y_i$ is the class of the $i^{th}$ sample. It is worth highlighting that the computation of the loss considers the entire batch, and the objective is to minimize the $\mathcal{L}_{D\mbox{-}loss}(X, f)$.

\subsection{Training and Weights Updating}

The training process using the D-loss is similar to training a traditional neural network.
The main difference relies on the fact that the last layer of the CNN model is the embedding layer instead of the commonly used $n$ neurons that represent the $n$ classes of the problem.
Still, the process of updating the weights is equal to the traditional.

Several batches are created from the training data. Their size is a hyper-parameter of the network and the selection of the data within a batch is random. 
The weights are updated after the processing of each mini-batch. First, a distance is calculated for each pair within a mini-batch, and, based on those distances, the genuine and impostor distribution curves are computed. Subsequently, the mean and standard deviation of curves are calculated to obtain the decidability. The learning process aims to minimize decidability. The entire process is presented in Algorithm~\ref{alg:dloss}.

\begin{algorithm}[!htb]
    \caption{D-loss on one mini-batch.}
    \label{alg:dloss}
    \SetAlgoNoLine%
    \textbf{Mini-Batch Setting:} The batch size $N$, the number of classes $C$.

    \textbf{Input:} $X = \{(x_i, y_i)\}^N_{i=1}$, the distance function $f$, the embedding function $d$, the learning rate $\beta$
    
    \textbf{Output: Updated $f$.}
    
    \textbf{Step 1:} Compute the embeddings by feeding-forward all images $\{x_i\}^N_{i=1}$.
    
    \textbf{Step 2:} Compute the pair-wise distances.
    
        \Indp
        \ForEach{$x_i \in X$}{
            \ForEach{$x_j \in X$}{
                pdist[i, j] = d($x_i$, $x_j$)
            }
        }
        \Indm
    
    \textbf{Step 3:} Compute the genuine and impostor scores:
        
        \Indp
        
        \ForEach{$x_i \in X$}{
            \ForEach{$x_j \in X$}{
                \If{$y_i$ == $y_j$ and $i$ != $j$}{
                    genuine.push(pdist[i, j])
                }
                \If{$y_i$ != $y_j$}{
                    impostor.push(pdist[i, j])
                }
            }
        }
        \Indm
    
    \textbf{Step 4:} Compute the decidability (Eq.~\eqref{eq:dec}).
    
    \textbf{Step 5:} Compute $\mathcal{L}_{D\mbox{-}loss}(X, f)$ (Eq.~\eqref{eq:dloss}).
    
    \textbf{Step 6:} Gradient computation and back-propagation to update the parameters of the network.
    
    \Indp$\nabla _f = \frac{\partial \mathcal{L}_{D\mbox{-}loss}(X, f)}{\partial f}$ ~~~~~~~~~~ and ~~~~~~~~~~ $f = e - \beta 	\cdot \nabla _f $
\end{algorithm}

\section{Experiments}
\label{sec:experiments}

This section presents the details of the experiments, the results, and the implications of the proposed methodology. Source code will be available after acceptance.

\subsection{Implementation Details}

We implement the CNN functions and loss function with the TensorFlow/Keras Library.
The CNN models are trained on a GPU GeForce Titan X with 12GB. In order to perform a fair comparison among losses and avoid experimental flaws such as pointed out by Musgrave \textit{et al.}~\cite{musgrave2020metric}, all losses are evaluated under the same conditions, fixing the network architecture, instances in the batches (same seed), optimization algorithm (Adam optimizer, initial learning rate of 10$^{-3}$), and same embedding dimension size.

\subsection{Datasets}

MNIST, Fashion-MNIST, and CIFAR-10 are used as proof of concept, mainly to investigate the convergence of D-loss in different domains. Also, these datasets are common benchmarks in machine learning.

We also employ a popular periocular benchmark to evaluate the proposed loss in a biometric scenario: CASIA-IrisV4. 
For a fair comparison, a 3-fold classification strategy is evaluated. 
All 249 individuals are split equally in three groups without overlap. 
All samples of an individual is just in one fold.
To conduct the experiments, we set 30\% of the train data for validation.

 \paragraph*{MNIST:} The MNIST dataset of handwritten digits~\citep{lecun2010mnist} consists of gray-scale images with a size of 28x28 each.
 There are ten classes (zero to nine), and each contains 6,000 images for training and 1,000 for testing. 

 \paragraph*{Fashion-MNIST:}
 Similar to MNIST, the Fashion-MNIST~\citep{xiao2017fashion} has gray-scale images of 28x28 resolution, describing fashion items (shoes, coat, etc.).  It is composed of 10 classes with the same MNIST distribution over the classes.

 \paragraph*{CIFAR-10:}
 The CIFAR-10~\citep{krizhevsky2009learning} is a 10 class problem, with 6,000 images per class. It is a collection of 60,000 colored images, in which 10,000 images are for testing and 50,000 for training. Each image is a RGB of size 32x32.

 \paragraph*{CASIA-IrisV4:} The CASIA-IrisV4 contains six subsets and is an extension of CASIA-IrisV3 dataset. In this work, the CASIA-Iris-Interval (subset from CASIA-IrisV3) is used to report the results. The iris images are from 249 different subjects with a total of 1,438 JPEG images captured under near infrared illumination.

\subsection{Evaluation Metrics}

Metrics such as precision, recall, and accuracy are not the most suitable ones for biometric learning problems.
Therefore, we use the False Acceptance Rate (FAR), False Rejection Rate (FRR), Equal Error Rate (EER), and Recall@K.

The EER is the point in which the FRR and FAR have the same values on the decision error trade-off curve. 
The FRR is the rate of incorrect rejections over different thresholds (zero to one), while the FAR is the rate of incorrect acceptances.
The EER is the point of intersection of both FAR and FRR.

\subsection{CNN Architecture and Model Training}

\begin{figure}[!htb]
    \centering
    \begin{tabular}{@{}c@{}}
        \includegraphics[width=0.9\linewidth,page=1]{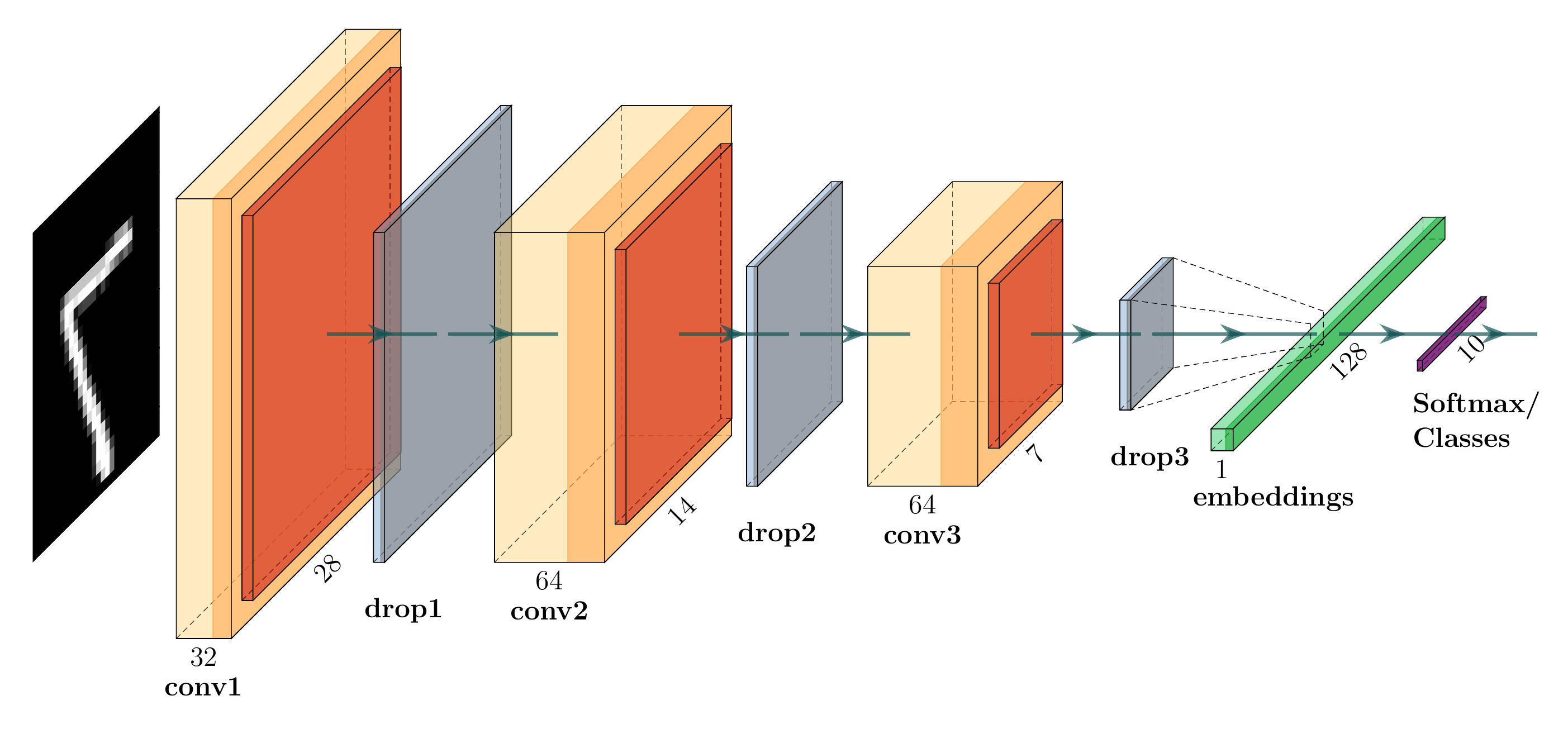} \\
            \parbox{.95\linewidth}{\footnotesize (a) MNIST-architecture proposed in which \textit{conv1}, \textit{conv2} and \textit{conv3} have filters size equal to 2x2, with \textit{ReLU} activation and padding to avoid reduction after the convolution. All \textit{poolings} are with the max function and window size of 2. \textit{Dropout} of 30\%.}\\
        
        \includegraphics[width=0.9\linewidth,page=2]{figs/cnns.pdf} \\
            \parbox{.95\linewidth}{\footnotesize (b) CIFAR-architecture proposed in which \textit{conv1}, \textit{conv2} and \textit{conv3} have filters size equal to 3x3, with \textit{ReLU} activation and padding to avoid reduction after the convolution. All \textit{poolings} are with the max function and window size of 2. \textit{Dropout} of 20\%.}
    \end{tabular}
    \vspace{-0.6em}
    \caption{Architectures used during experiments.}
    \label{fig:nets}
\end{figure}

We run our experiments on three different architectures.
Two architectures are small networks (Figure~\ref{fig:nets}) and each one matches the input size resolution of MNIST and CIFAR datasets ($28 \times 28$ and $32 \times 32$). The architectures used with the MNIST, Fashion, and CIFAR-10 make use of simple operations that are well understood in the literature. The rationale to use such architectures (with simple convolutional blocks) is to emphasize the impact of the loss function.

An architecture with more capacity is selected for the CASIA-IrisV4 dataset. The experiments are run on the EfficientNet family network (B0 model) \citep{tan2019efficientnet}. The architecture is larger, more sophisticated and has an input size resolution of $224 \times 224$.

All architectures have an embedding layer with dimension 256 (256-feature array output) with L2 normalization. The MNIST-architecture has 100,010 parameters, the CIFAR-architecture 567,082 parameters, and CASIA-V4-architecture 4,377,500 parameters. Before the loss layer, a Lambda Layer with an L2 normalization is used. 

With these three architectures, we are able to evaluate the D-loss in different scenarios and from shallow and simple networks to one state-of-the-art network such as EfficientNet B0.

No data augmentation is implemented during training, and the euclidean distance is used to calculate the dissimilarity scores.

We employ Adam optimizer to train the models with a learning rate of $10^{-3}$, and all images normalized to the range $[0,1]$. The number of epochs is the same for all experiments, and dataset dependent: $100$ epochs for the MNIST, $500$ epochs for the Fashion-MNIST, $2,000$ epochs for CIFAR-10 and 1,000 epochs for CASIA-V4.

An initial experiment is performed on the MNIST dataset, to assess the impact of the batch size on both D-loss and Triplet-loss. We evaluate the batch sizes: 300, 400, and 500.
As shown in Table~\ref{tab:batch}, the batch size equal to 400 performed better on validation data for the D-loss. For triplet loss, results are very similar, both for batch sizes 300 and 400. Thus, a batch size of 400 is fixed for all losses to maintain the same evaluation setup in all experiments related to MNIST, Fashion and CIFAR-10. For the experiments on the CASIA-V4, a batch size of 150 is employed due to the network architecture size and hardware constraints.

\begin{table}[!htb]
    \caption{Batch size investigation on MNIST Dataset. Results on validation set (30\% of train set).} 
    \label{tab:batch}
    \centering
    \begin{tabular}{c|c|c|c|c}
        \toprule\toprule
        \multirow{2}{*}{Dataset} 
            & \multicolumn{2}{c|}{\parbox{3.3cm}{\centering D-loss}}
            & \multicolumn{2}{c}{\parbox{3.3cm}{\centering Triplets Soft-hard Loss}}\\  \cmidrule{2-5}
                & \parbox{1.5cm}{\centering EER (\%)} & Dec
                & \parbox{1.5cm}{\centering EER (\%)} & Dec \\
                \hline
        300 & 1.07 & 7.92 & 0.70 & 7.63 \\
        400 & 0.47 & 9.31 & 0.74 & 7.25 \\
        500 & 1.01 & 8.22 & 0.83 & 7.10 \\
        \bottomrule\bottomrule
    \end{tabular}
\end{table}

\subsection{Results and Literature Comparison}

Tables~\ref{tab:results:eer} and~\ref{tab:results:recall} present a comparison between the D-loss, the Triplets soft-hard loss, the Multi Similarity Loss, and the Softmax-loss. The proposed approach outperformed the others in two out of four evaluated scenarios. On the remaining it exhibit comparable performance.

\begin{table}[!htb]
    \caption{Reported results in terms of Equal Error Rate (EER) in \%. MS = Multi-Similarity; TS = Triplets Soft-hard; * = Model did not converge; $^\star$ = No statistical significance.}
    \label{tab:results:eer}
    \centering
    \begin{tabular}{c|c|c|c|c}
        \toprule\toprule
        Dataset      & D-loss       & MS Loss      & TS Loss* & Softmax \\  \hline
        MNIST (10)   &  1.04 &  1.06 &  \best{0.91} &  1.50 \\
        Fashion (10) &  \best{5.38} &  5.82 &  5.94 &  7.58 \\
        CIFAR-10     & \best{13.01} & 14.11 & 20.01 & 14.08 \\
        CASIA-V4     & \best{7.96 $\pm$ 3.43$^\star$} & \best{7.84 $\pm$ 1.29$^\star$} & 38.55 $\pm$ 53.25 & 9.4 $\pm$ 1.92 \\
        \bottomrule\bottomrule
    \end{tabular}
\end{table}

\begin{table}[!htb]
    \caption{Reported results in terms of Recall@K (R@K). MS = Multi-Similarity; TS = Triplets Soft-hard; * = Model did not converged.}
    \label{tab:results:recall}
    \centering
    \resizebox{0.95\textwidth}{!}{    
    \begin{tabular}{c|c|c|c|c|c}
        \toprule\toprule
        Dataset      & R@K & D-loss      & MS Loss     & TS Loss* & Softmax \\  \hline
        \multirow{4}{*}{MNIST (10)}
            & 1 & 0.99 & 0.98 & 0.99 & 0.99 \\
            & 2 & 0.99 & 0.99 & 0.99 & 1.00 \\
            & 4 & 0.99 & 0.99 & 1.00 & 1.00 \\
            & 8 & 0.99 & 1.00 & 1.00 & 1.00 \\ \hline
        \multirow{4}{*}{Fashion (10) }
            & 1 & 0.88 & 0.88 & 0.89 & 0.89 \\
            & 2 & 0.93 & 0.93 & 0.94 & 0.94 \\
            & 4 & 0.96 & 0.96 & 0.96 & 0.96 \\
            & 8 & 0.97 & 0.98 & 0.97 & 0.98 \\ \hline
        \multirow{4}{*}{CIFAR-10}
            & 1 & 0.77 & 0.77 & 0.79 & 0.74 \\
            & 2 & 0.84 & 0.84 & 0.85 & 0.82 \\
            & 4 & 0.89 & 0.89 & 0.90 & 0.88 \\
            & 8 & 0.92 & 0.92 & 0.93 & 0.92 \\ \hline
        \multirow{5}{*}{CASIA-V4}
            &  1 & 0.99 $\pm$ 0.01 & 0.95 $\pm$ 0.03 & 0.66 $\pm$ 0.57 & 0.99 $\pm$ 0.01 \\
            &  2 & 0.99 $\pm$ 0.00 & 0.98 $\pm$ 0.01 & 0.67 $\pm$ 0.58 & 0.99 $\pm$ 0.01 \\
            &  4 & 1.00 $\pm$ 0.01 & 0.99 $\pm$ 0.01 & 0.67 $\pm$ 0.58 & 0.99 $\pm$ 0.01 \\
            &  8 & 1.00 $\pm$ 0.01 & 1.00 $\pm$ 0.00 & 0.67 $\pm$ 0.58 & 1.00 $\pm$ 0.00 \\
            & 16 & 1.00 $\pm$ 0.01 & 1.00 $\pm$ 0.00 & 0.67 $\pm$ 0.58 & 1.00 $\pm$ 0.00 \\
            & 32 & 1.00 $\pm$ 0.00 & 1.00 $\pm$ 0.00 & 0.67 $\pm$ 0.58 & 1.00 $\pm$ 0.00 \\
        \bottomrule\bottomrule
    \end{tabular}
    }
\end{table}

All scenarios present training overfitting as well. However, for the simple datasets (MNIST and Fashion-MNIST), a balanced degree of specialization and generalization is observed for all three losses.

\subsection{Discussion}

The main advantage of the D-loss over the triplets is that it does not require hard samples selection, which is a costly and difficult operation. That operation substantially increases the complexity of triplet-based losses. 

A considerable disadvantage of the proposed loss is related to batch size. While the majority of the pair-based distance losses need an anchor and at least a negative sample, the proposed approach depends on a large number of both positive and negative samples to create the similarity (or dissimilarity) score distribution curves for the loss computation.

Regarding memory consumption, all pair-based losses analyzed here are comparable.
While the D-loss and the triplets store distances between pairs, the multi-similarity loss stores the multiplication of the embeddings.

\begin{figure}[!htb]
    \centering
    \begin{tabular}{c}
        \includegraphics[width=.85\linewidth,trim={0cm 0cm 0cm 1.2cm}, clip]{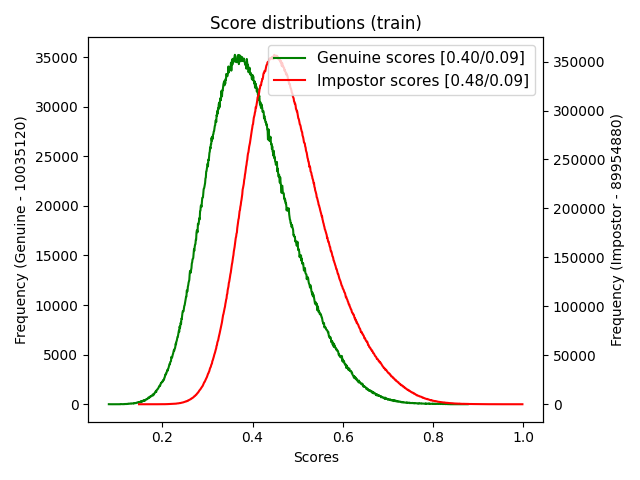} \\
        (a) \\
        \includegraphics[width=.85\linewidth,trim={0cm 0cm 0cm .80cm}, clip]{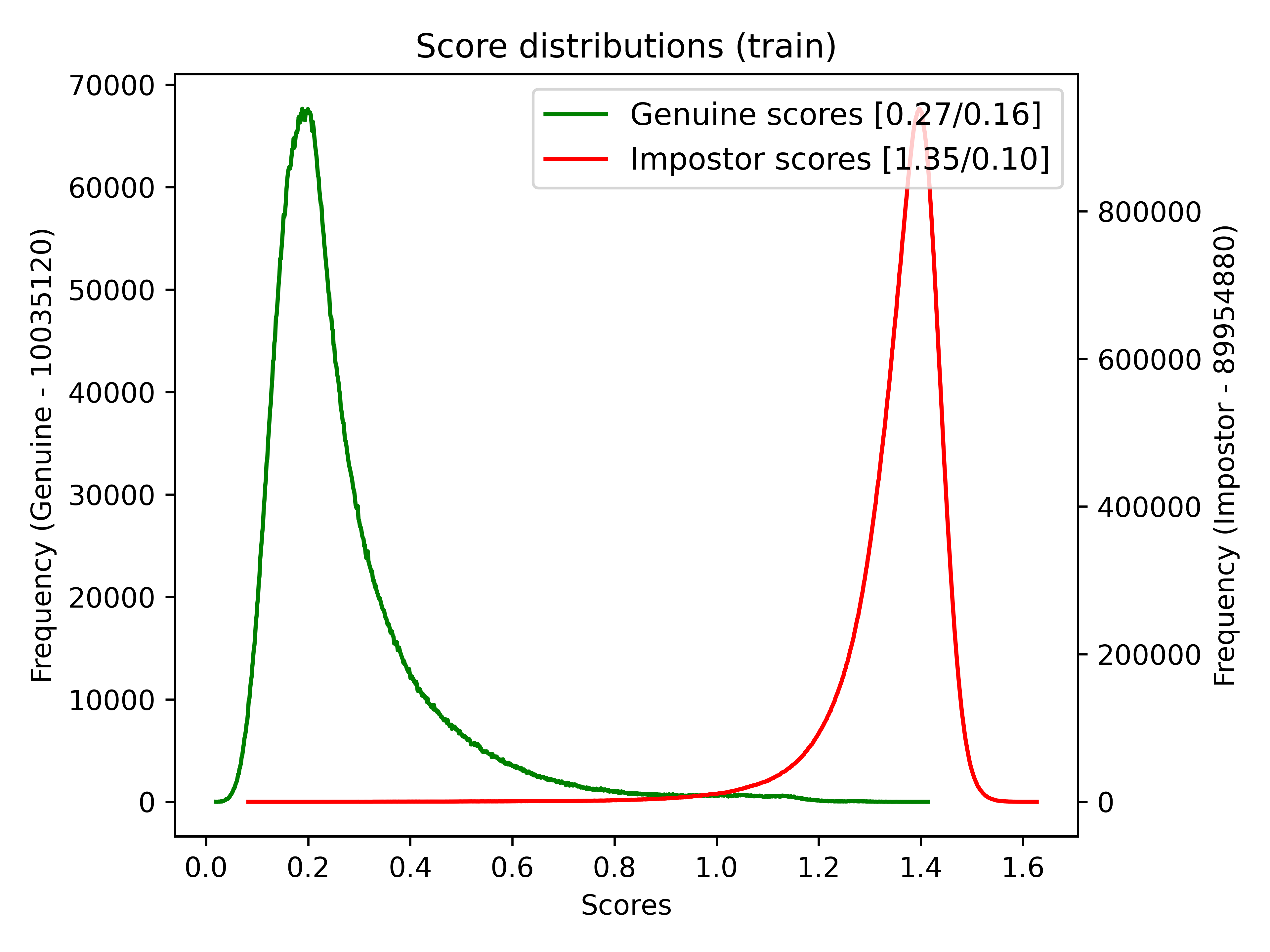} \\
        (b)
    \end{tabular}
    \vspace{-0.6em}
    \caption{Distribution scores before (a) and after (b) training the model with D-loss on MNIST dataset. More distance between the mean of the curves implies better inter-class discrimination and the reduction in the width of the genuine distribution curve implies intra-class improvement.}
    \label{fig:curves}
\end{figure}

The impact of training with D-loss is more apparent in the genuine and impostor distributions curves, as one can see in Figure~\ref{fig:curves}.
Figure~\ref{fig:curves} (a) corresponds to the non-trained model, and Figure~\ref{fig:curves} (b) is the result after training.

\section{Conclusion}
\label{sec:conclusion}

Based on the decidability index, which is intuitive, easy to compute, and implement, we propose the D-loss as an alternative to triplet-based losses and the softmax-loss.
The D-loss function is more suitable than softmax-loss for training models aiming to feature extraction / data representation, much like the Triplet-base loss.
Moreover, the D-loss avoids some Triplet-based loss disadvantages, such as the use of hard samples, and it is non-parametric. Also, triplet-based losses have tricky parameter tuning, which can lead to slow convergence.
The D-loss drawback is related to memory consumption since it requires a large batch size which can hinder the use of larger models.

The D-loss surpassed three other popular loss functions (Triplets Soft-hard Loss, Multi-Similarity Loss, and Softmax-Loss) in two out of three popular benchmark problems (MNIST-FASHION and CIFAR-10) and presented comparable results on a challenging scenario with the CASIA-IrisV4 dataset.

\section*{Conflict of interest statement}

The authors declare that they have no known competing financial interests or personal relationships that could have appeared to influence the work reported in this paper.


\bibliography{main}

\end{document}